\documentclass[journal]{IEEEtran}

\usepackage{xcolor,soul,framed} 

\colorlet{shadecolor}{yellow}
\usepackage[pdftex]{graphicx}
\graphicspath{{../pdf/}{../jpeg/}}
\DeclareGraphicsExtensions{.pdf,.jpeg,.png}

\usepackage[cmex10]{amsmath}
\usepackage{array}
\usepackage{mdwmath}
\usepackage{mdwtab}
\usepackage{eqparbox}
\usepackage{url}

\usepackage{amssymb}
\usepackage{multirow}
\newcommand{\RNum}[1]{\uppercase\expandafter{\romannumeral #1\relax}}
\usepackage[justification=centering]{caption}

\begin{document}
\bstctlcite{IEEEexample:BSTcontrol}
    \title{Multi-scale temporal network for continuous sign language recognition}
  \author{Qidan~Zhu,
      Jing~Li*,
      Fei~Yuan,
      Quan~Gan

 \thanks{*Corresponding author\par
Email addresses: \par
zhuqidan@hrbeu.edu.cn (Qidan Zhu), \par
ljing@hrbeu.edu.cn (Jing Li), \par
bohelion@hrbeu.edu.cn (Fei Yuan), \par
gquan@hrbeu.edu.cn (Quan Gan)}
  }

\maketitle

\begin{abstract}
Continuous Sign Language Recognition (CSLR) is a challenging research task due to the lack of accurate annotation on the temporal sequence of sign language data. The recent popular usage is a hybrid model based on “CNN + RNN” for CSLR. However, when extracting temporal features in these works, most of the methods using a fixed temporal receptive field and cannot extract the temporal features well for each sign language word. In order to obtain more accurate temporal features, this paper proposes a multi-scale temporal network (MSTNet). The network mainly consists of three parts. The Resnet and two fully connected (FC) layers constitute the frame-wise feature extraction part. The time-wise feature extraction part performs temporal feature learning by first extracting temporal receptive field features of different scales using the proposed multi-scale temporal block (MST-block) to improve the temporal modeling capability, and then further encoding the temporal features of different scales by the transformers module to obtain more accurate temporal features. Finally, the proposed multi-level Connectionist Temporal Classification (CTC) loss part is used for training to obtain recognition results. The multi-level CTC loss enables better learning and updating of the shallow network parameters in CNN, and the method has no parameter increase and can be flexibly embedded in other models. Experimental results on two publicly available datasets demonstrate that our method can effectively extract sign language features in an end-to-end manner without any prior knowledge, improving the accuracy of CSLR and achieving competitive results.
\end{abstract}

\begin{IEEEkeywords}
Continuous Sign Language Recognition; Multi-scale Temporal Features; Temporal Receptive Field; Multi-level CTC loss
\end{IEEEkeywords}

%
\IEEEpeerreviewmaketitle


\section{Introduction}

\IEEEPARstart{S}{ign} language conveys semantic information through hand movements, gesture appearance, etc. It is the main way of communication between deaf people or between deaf people and normal people. With the increasing number of hearing-impaired people worldwide, sign language recognition occupies an increasingly important position, involving various fields such as computer vision, natural language processing and human-computer interaction technologies, and has gained extensive attention\cite{cui2019deep}\cite{huang2021boundary}\cite{koller2020quantitative}. Video-based sign language recognition is divided into two categories: one is isolated word recognition, where each video segment represents only a single sign language word\cite{pan2020attention}\cite{hu2021global}, and the other is CSLR, where each video segment represents a sign language sentence\cite{xiao2020multi}\cite{hassan2019multiple}\cite{choudhury2017movement}. For real-life applications, CSLR research is of more social value.\par

At present, CNN is becoming more and more popular in CSLR research due to its powerful feature representation ability and sequence modeling capability. A series of deep learning-based CSLR models continue to emerge and perform well\cite{sharma2021continuous}\cite{koller2018deep}\cite{wei2020semantic}. It is mainly based on CNN to extract frame-wise features and time-wise features of video clips. Gloss is the basic unit in CSLR. Video-based sign language recognition is to translate the video sign language action sequence into a continuous gloss, which in turn is translated into natural language to help hearing impaired people communicate with others\cite{guo2018hierarchical}. Due to the lack of strict correspondence between video frames and annotation sequences, that is, only the sign language actions and sequences performed by the sign language presenter in the video are known, but the start and end moments of the specific actions are unknown, so CSLR is a weakly supervised learning problem\cite{koller2017re}. This places a high demand on sequence learning, that is, it is crucial to learn the correspondence between video sequences and labeled lexical sequences.\par

\begin{figure*}
  \begin{center}
  \includegraphics[width=5in]{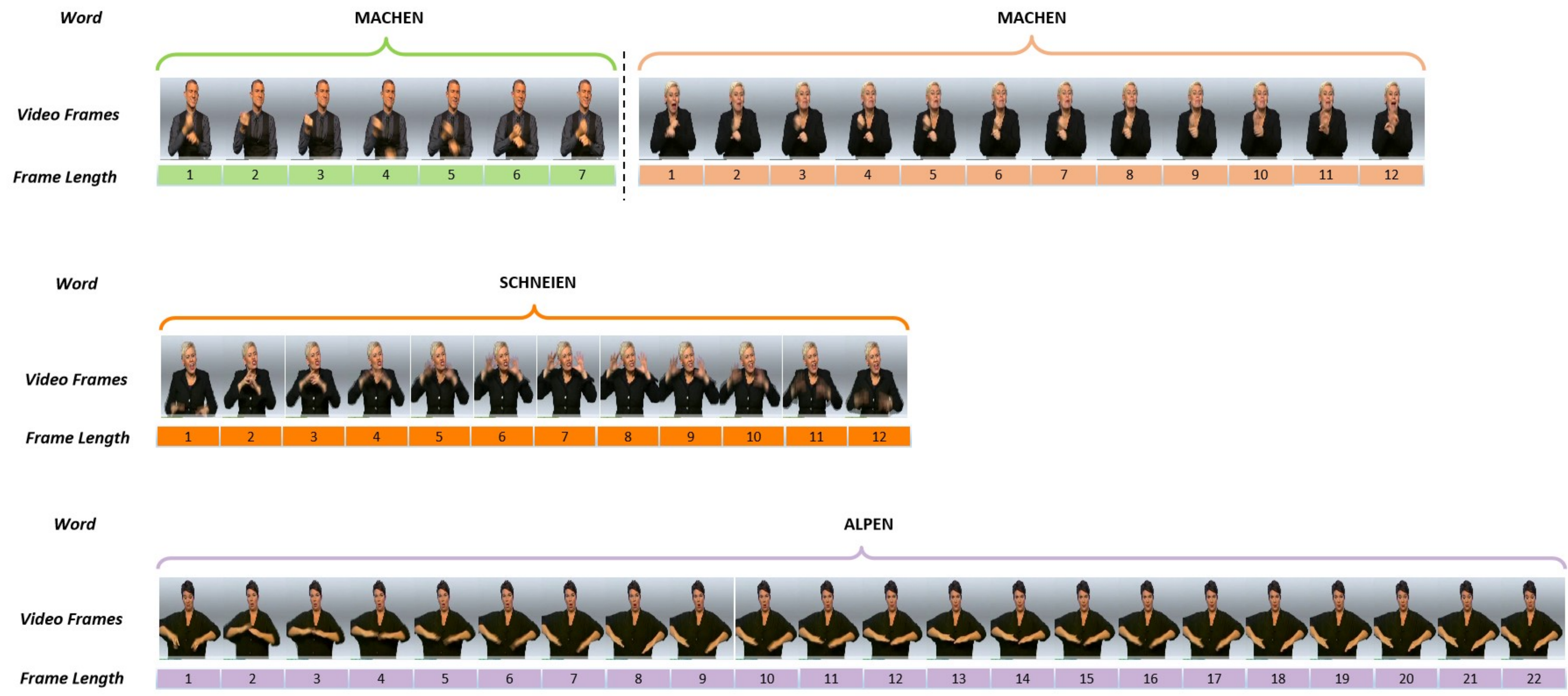}
  \caption{The video frame lengths for different glosses and for the same gloss demonstrated by different people (the first row of images shows the respective video frame lengths for the same gloss demonstrated by different people, the second and third rows of images show the video frame lengths for different glosses).}\label{fig:ljxy1}
  \end{center}
\end{figure*}

Current sequence learning models include 3D-CNN\cite{hao2022spatiotemporal}, RNN\cite{zhao2021real}, Long Short-Term Memory (LSTM)\cite{huang2021lstm}, etc. 3D-CNNs have a strong representation ability for videos and have achieved excellent performance on video action recognition tasks. The 3D convolution operation is able to model not only the spatial information but also consider the temporal information between video frames. However, it suffers from the problem of bulky model and large number of parameters. RNN has the ability to obtain the information of the input sequence and output the sequence, which is widely used in sequence recognition, but its ability to acquire long-distance dependency information is weak. The proposal of LSTM alleviates this problem, but this problem still exists, and the convergence speed is slow. 1DCNN is widely used in CSLR for temporal feature extraction in recent years because of its advantages of simple structure and small number of parameters\cite{wei2020semantic}\cite{xie2021multi}\cite{min2021visual}. Most of the CSLR models based on these algorithms in recent years have been extracted from local features of fixed temporal receptive fields\cite{min2021visual}\cite{cheng2020fully}. However, in the original video sequence, the length of the video clip sequence corresponding to different glosses is different, and the proficiency of sign language and some other interference caused by different sign language performers in the process of presentation will cause the same gloss to have inconsistent time used, as in Figure 1. In this case, the results obtained by using the fixed temporal receptive field for feature extraction will also be different, resulting in the limitations of the extracted time-wise features and affecting the final recognition performance. Meanwhile, CTC\cite{li2020reinterpreting} is widely used for decoding sequences in CSLR because of its ability to optimize the neural network directly without segmentation and annotation for unsegmented sequence data. However, using a single CTC loss for deep neural networks trained by a backward propagation algorithm based on the chain rule causes the shallow network parameters to be poorly learned and updated, which can affect the model fit.\par

To address the above issues, this paper proposes an end-to-end CSLR network, and it is more in line with the real scene. The network mainly consists of three parts. For the output of the frame-wise feature extraction part composed of Resnet and two fully connected layers, the proposed time-wise feature extraction part is used for temporal feature learning. Specifically, the proposed MST-block is first used to extract different scale temporal receptive field features and combine them into a candidate space, which is fused by learnable parameters, and then further encoded by the transformers module to obtain more accurate temporal features. Finally, the proposed multi-level CTC loss part is used for training to obtain recognition results. The research in this paper is based on 1DCNN, using temporal receptive fields of different scales to enhance sequence modeling capability, followed by the self-attention mechanism of transfomers\cite{mazzia2022action}\cite{xie2021pisltrc} to better access long-distance dependent information and improve the discriminative power of features. Finally using the proposed multi-level CTC loss, which not only can better decode the temporal features, but also can make the parameters of the shallow network well updated, and then efficiently train the frame-wise feature extraction network and temporal modeling network, further improving the recognition performance. A similar work to this paper is the VAC network proposed by Min et al.\cite{min2021visual}, where both we and that paper use CNN for frame-wise feature extraction, 1DCNN for time-wise feature extraction, and finally CTC loss for training recognition decoding. The differences between us and them are: 1) We propose a multi-scale temporal module when using 1DCNN for time-wise feature extraction, which improves the accuracy of the extracted temporal features compared to the 1DCNN with fixed receptive field they use; 2) The final visual alignment module in that paper uses two different levels of CTC loss and a self-distillation loss, on the basis of which in this paper, a multi-level CTC loss is proposed to further enhance the training of the shallow layers in a deep network. \par

The main contributions of this paper are as follows:\par

\begin{enumerate}
\item The proposed end-to-end model MSTNet improves the accuracy of CSLR model on two publicly available datasets and achieve competitive results.\par
\item The proposed MST-block better exploits the temporal receptive fields of different scales, and has a significant improvement in the final recognition performance.\par
\item The proposed multi-level CTC loss trains the model, which enables the shallow network parameters to be better learned and updated, and has a large improvement on the experimental results in the final experiments. The method has no parameter increase and can be flexibly embedded into other models.\par
\end{enumerate}

\begin{figure*}
  \begin{center}
  \includegraphics[width=5in]{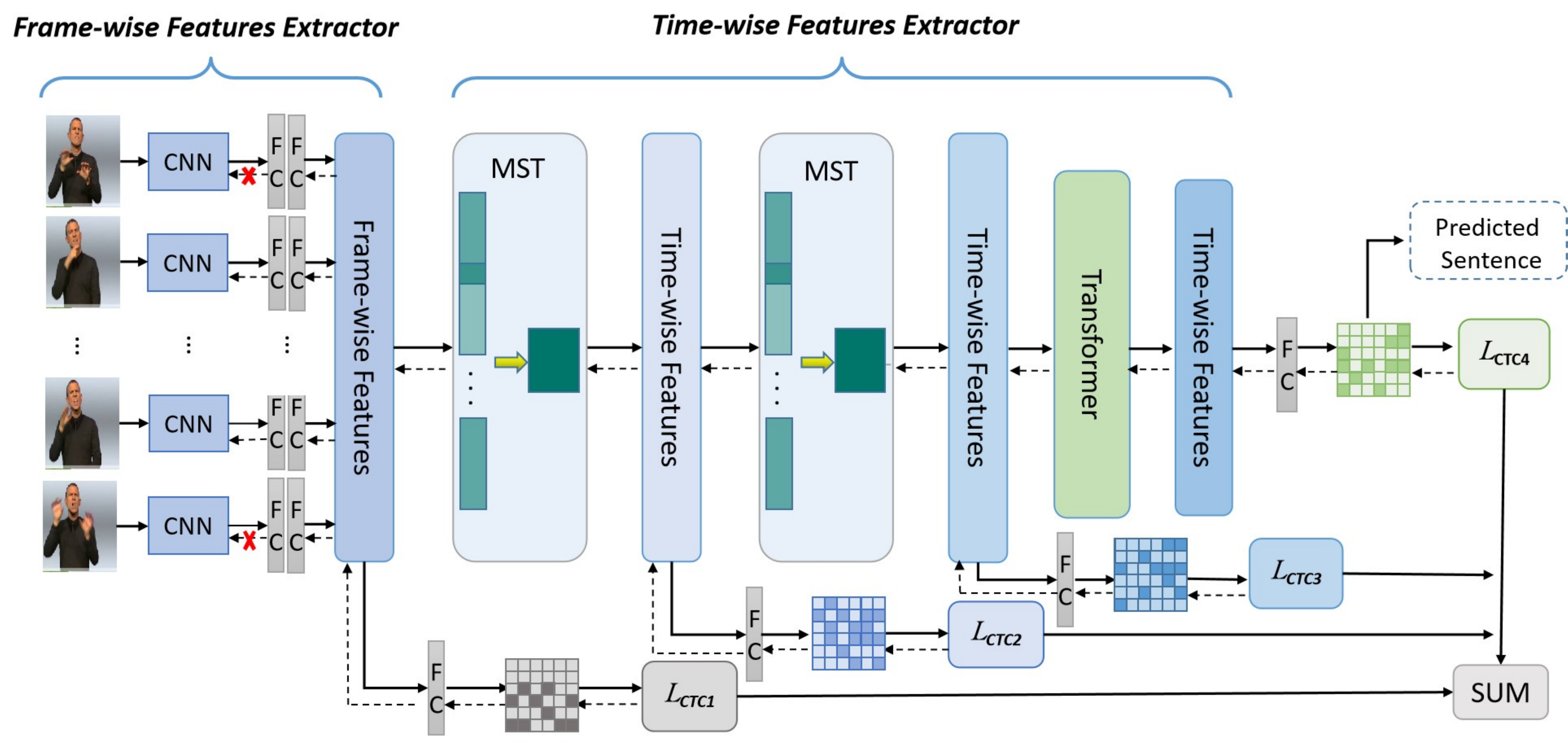}
  \caption{The overall framework of the model proposed in this paper.}\label{fig:ljxy2}
  \end{center}
\end{figure*}

\section{Related work}

The mainstream models for CSLR mainly consist of frame-level feature extraction, temporal feature modeling and finally decoding recognition. In recent years, with the rapid development of deep learning, the network structure of “CNN+LSTM/Bi-LSTM+CTC” is mostly used for CSLR.\par

First of all, the frame-wise feature extraction of sign language is basically based on CNN. Many classic CNN network structures that have been successfully applied in computer vision also play an important role in sign language frame-wise feature extraction. Koller et al\cite{koller2017re}. used GoogLeNet for frame-wise feature extraction; Cui et al.\cite{cui2019deep} used VGG-S and GoogLeNet for frame-wise feature extraction for hand-shaped sequences and full image sequences, respectively; Min et al.\cite{min2021visual} used ResNet18 performs frame-wise feature extraction. Cheng et al.\cite{cheng2020fully} self-constructed a CNN network for feature extraction. Although GoogLeNet is used as a feature extraction model in many papers, its model is complex and the number of parameters is too large. Considering the great success of Resnet in the field of image recognition, this paper adopts Resnet for frame-wise feature extraction.\par

Then, since CSLR is weakly supervised learning, it is particularly important to learn the correspondence between sequential representations and lexical labels, that is, temporal features play a very important role in CSLR. A series of algorithm models on sequence learning have been continuously proposed to improve the temporal feature modeling capability. Koller et al.\cite{koller2019weakly} proposed a new approach to sequence learning, exploiting sequence constraints in each independent data stream and combining them by explicitly imposing synchronization points to exploit the parallelism shared by all sub-problems, according to the hybrid approach embeds a powerful LSTM model in each HMM stream for recognition. Cui et al.\cite{cui2019deep} developed a CSLR framework with a deep neural network and obtained the best results at that time, in which the sequence learning module used “1DCNN+BiLSTM” , their proposed iterative alignment optimization and multimodality both contribute to the accuracy improvement. Recent models such as the FCN network proposed by Cheng\cite{cheng2020fully} and the VAC network proposed by Min\cite{min2021visual} have used 1DCNN for temporal feature extraction and achieved good results. These algorithms have improved the acquisition of time-wise features to a certain extent, but they all extract the local features for a fixed temporal receptive field. The research of Wei et al.\cite{wei2020semantic} uses pooling of different strides for downsampling for temporal features to generate temporal features of different scales. However, this method is multi-scale in the time series length, and there is no information fusion between temporal features of different scales, and the pooling itself will lose a certain amount of information. This paper proposes a MST-block using different scales of time receptive fields to improve temporal modeling capabilities. Our MST-block is multi-scale on the convolution kernel, using convolution kernels of different scales to calculate temporal features in parallel and then using learnable parameters for feature fusion and output. The temporal features obtained by our method will be richer and more comprehensive.\par

Finally, for the process of decoding the vectors encoded with temporal features into sequences for recognition, CTC loss is the most popular algorithm nowadays. CTC can handle unsegmented input sequence data and help the model learn the correspondence between input and output sequences. Yang et al.\cite{yang2019sf} proposed a structured feature network (SF-Net), extract features in a structured way, and gradually encode frame-level, gloss-level, and sentence-level information into feature representations, and finally decode them using CTC loss. The architecture proposed by Koishybay et al.\cite{koishybay2021continuous} involves spatiotemporal feature extraction model to segment useful "gloss" features and BiLSTM with CTC as sequence model. The use of CTC in these models has achieved good results, but basically a single CTC loss is directly used in the final recognition part. Models such as the model proposed by Min\cite{min2021visual} and the model proposed by Wei\cite{wei2020semantic} do not use a single CTC loss. Although they all improve the performance of the model to some extent, there is still potential for further improvement. Because the deep neural network uses the back propagation algorithm based on the chain rule to train the network, which will cause the shallow network parameters to be poorly learned and updated to a certain extent. In this paper, a multi-level CTC loss is proposed in this paper, which not only can better decode the temporal features, but also can make the parameters of the shallow network well updated, and then efficiently train the frame-wise feature extraction network and temporal modeling network, further improving CSLR performance.\par

\begin{figure*}
  \begin{center}
  \includegraphics[width=5in]{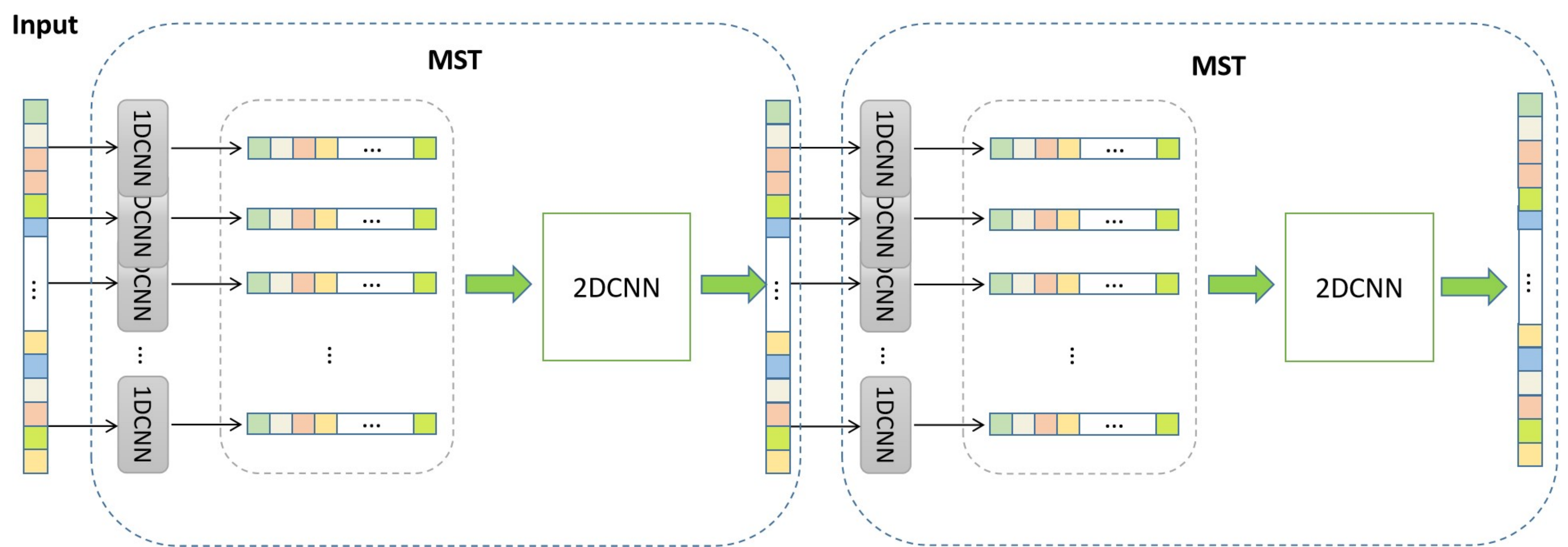}\\
  \caption{Details of the multi-scale temporal block(two MST-bolocks are used in this paper, and the timing length is down-sampled for each passing MST-bolock).}\label{fig:ljxy3}
  \end{center}
\end{figure*}

\section{Approach}

The overall framework of the end-to-end CSLR proposed in this paper is shown in Figure 2. The model consists of three parts: frame-wise feature extraction part, time-wise feature extraction part and multi-level CTC loss training part. There are four levels of gloss features defined in our model. For the input sign language video, the frame-wise features, i.e., the first-level gloss features, are first obtained by the Resnet and two FC layers. After that, the time-wise features are obtained by the proposed MST-block. Specifically, the frame-wise features are passed through the first MST-block to get the second-level gloss features, and the second-level gloss features are sequentially used as the input of the second MST-block to get the third-level gloss features, which are then defined as the fourth-level gloss features after transfomers timing coding, i.e., the final timing features. Finally, for the four gloss features obtained, they are summed and trained to optimize model using multi-level CTC loss, and the final sign language recognition results are obtained using the fourth-level gloss features. The details of the proposed MST-block will be described in detail in section 3.2\par

\subsection{Frame-wise feature extraction}

The frame-wise feature extraction consists of a main network feature extractor and two fully connected layers. For an input sign language video $V=(x_1,x_2,...,x_T)=\{{x_t|_1^T\in \mathbb{R}^{T\times c\times h\times w}}\}$ 
containing T frames, where $x_t$ is the t-th frame image in the video, $h*w$ is the size of $x_t$, c is the number of channels, here $c=3$ for RGB video. V is input into the Resnet feature extractor $F_r$ to obtain feature expression $f_1=F_r(V)\in \mathbb{R}^{T\times c_1}$, and then after two fully connected layers to obtain feature expression $f_2=F_{fc}(f_1)\in \mathbb{R}^{T\times c_2}$, which is the final frame-wise feature vector containing spatial information with fixed dimensions. We define it as a first-level gloss feature in this paper. The sizes of $c_1$ and $c_2$ here are 512 and 1024, respectively. The function of adding two fully connected layers after the main network extracts features is to integrate the features in the image feature maps that have passed through multiple convolutional layers and pooling layers to obtain the high-level meaning of the image features. For the consideration of GPU memory, stochastic gradient stopping\cite{niu2020stochastic} is used between the feature extraction of the main network and the fully connected layer to reduce the memory usage and improve the training speed.\par

\subsection{Time-wise feature extraction}

For the frame-wise features extracted in section A, sequence modeling is performed. Here, the time-wise feature extraction part proposed in this paper includes: MST-block and transformers. After MST-block, the second-level gloss features and the third-level gloss features are obtained respectively, and finally the fourth-level gloss features are obtained after transfomers.\par

\begin{figure*}
  \begin{center}
  \includegraphics[width=2in]{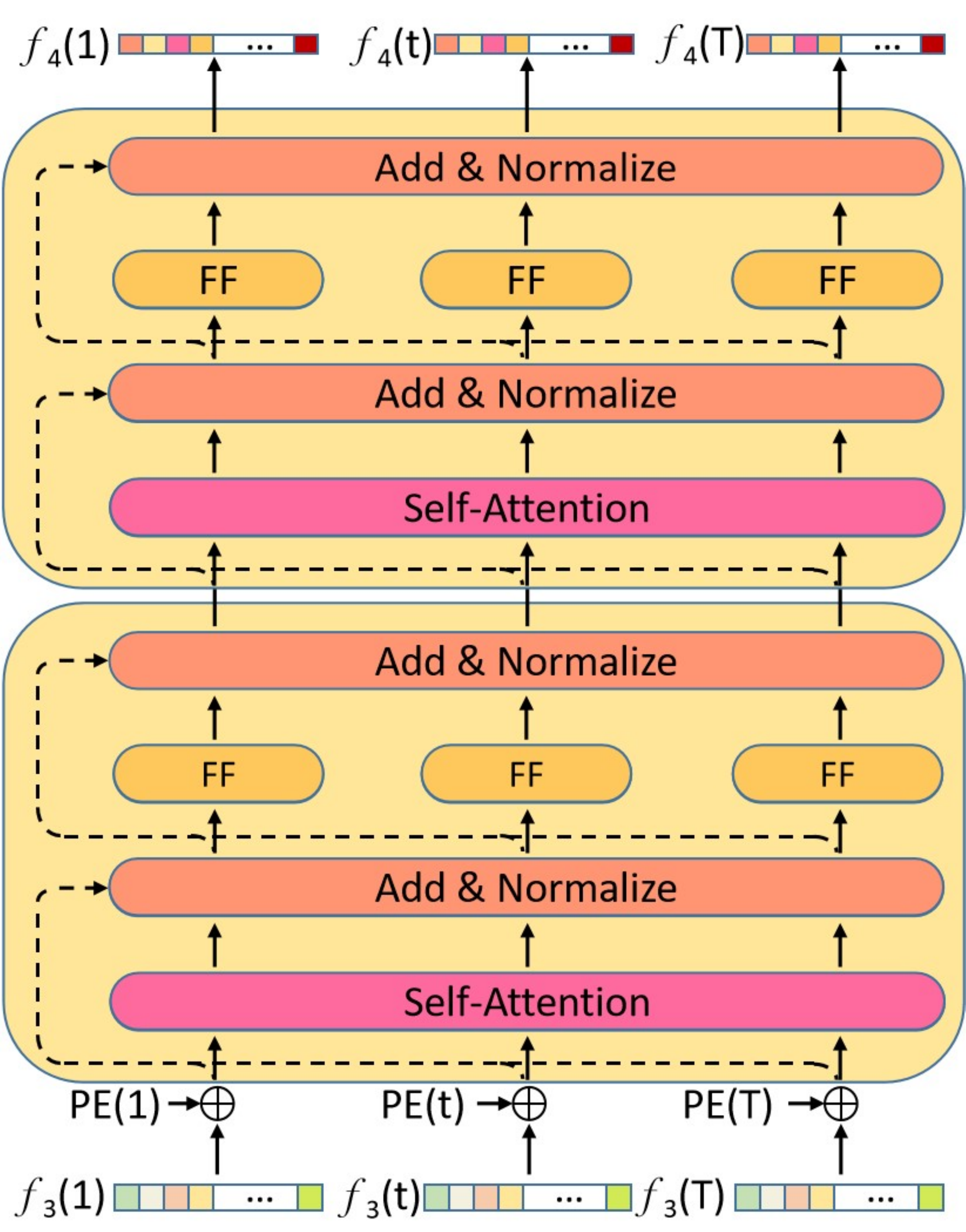}
  \caption{Details of the transformers encoding module (this paper uses the two-layer transformers encoding module to further encode the temporal features).}\label{fig:ljxy4}
  \end{center}
\end{figure*}

\subsubsection{MST-block}

In recent years, excellent CSLR models have emerged one after another, but most of them extract local features of fixed temporal receptive fields. In the real scene, the sequence lengths of the video clips corresponding to different glosses in CSLR are different, and the proficiency of sign language by different sign language performers and some other disturbances during the demonstration process cause inconsistency in the time used for the same gloss. In this case, the results obtained by using the fixed temporal receptive field for feature extraction will also be different, which will affect the performance of temporal modeling. In this paper, we first propose MST-block, as shown in Figure 3, which uses different scales of temporal receptive fields to improve the capability of temporal modeling.\par

The MST-block is mainly composed of multi-scale feature extraction and feature fusion. Multiple 1DCNNs with different convolution kernels are connected in parallel to form a multi-scale feature extraction part. The formula of 1DCNN is as follows:\par

\begin{equation}
y(t) = f_2(t)\times w(t)=\sum_{\substack{K}}^{\substack{i=0}}f_2(t-i)w(i)
\end{equation}

Where $f_2\in \mathbb{R}^{T\times c_2}$ is the weight, $f_2\in \mathbb{R}^{T\times c_2}$ is the output data, K is the size of the convolution kernel, $f_2\in \mathbb{R}^{T\times c_2}$ and T is the timing length.\par

For the first-level gloss feature obtained in 3.1, the dimension of the feature is first transformed, that is, $f_2\in \mathbb{R}^{T\times c_2}$ becomes $f_2^T\in \mathbb{R}^{c_2\times T}$. After that, it goes through the multi-scale feature extraction part. The multi-scale 1DCNN convolutional kernels have different sizes and the same number of channel dimensions, and the size of the timing and the number of features does not change during the feature extraction process. The size of the first convolution kernel is 3, the maximum convolution kernel size is M, and the step size is 2, that is, the size of each convolution kernel is increased by 2 on the size of the previous convolution kernel.\par

\begin{equation}
f_2^{'}=cat(y_n(t))
\end{equation}

Where $f_2^{'}$ is the output after multi-scale 1DCNN and n is the number of 1DCNNs.\par

After a 2DCNN for feature fusion and down-sampling by 2 times, $f_3\in \mathbb{R}^{c_2\times T_1}$ is obtained, that is, the second-level gloss feature, where $T_1=\frac{T}{2}$, repeat this process to obtain the third-level gloss feature $f_4\in \mathbb{R}^{c_2\times T_2}$, $T_2=\frac{T_1}{2}$.\par

\subsubsection{Transformers encoding}

Transformers model is a classical model of natural language processing (NLP) proposed by a team at Google in 2017, which uses the Self-Attention mechanism and does not use the sequential structure of RNNs, allowing the model to be trained in parallel and to have global information. The transformers encoding module used in this paper is shown in Figure 4. For the temporal feature vector obtained by MST-block, the temporal sequence is further encoded using the transformers encoding module, which results in more accurate temporal features.\par

Two identical transformers encoding modules are used in our proposed model. The Multi head Self Attention (MHSA) and the fully connected feedforward (FF) part constitute the transformers encoding module. The third-level gloss feature $f_4\in \mathbb{R}^{c_2\times T_2}$ plus the corresponding position information PE is used as the input of MHSA, and then through the temporal feature $f_4^{'}\in \mathbb{R}^{c_2\times T_2}$ obtained by FF, the same operation is repeated to obtain the final temporal feature $f_5\in \mathbb{R}^{c_2\times T_2}$ , that is, the fourth-level gloss feature. MHSA not only extends the ability of the model to focus on different positions, but also enhances the ability of the attention mechanism to express the roles between words within the sentences of interest. Compared with single-head self-attention, each head in MHSA maintains a Q, K, V matrix of its own to achieve different linear transformations, so that each head also has its own special expressive information. FF, on the other hand, strengthens the representation in a non-linear way, making the features more expressive.\par

\subsubsection{Multi-level CTC loss}
CSLR belongs to weakly supervised learning. The input video is an unsegmented sequence and lacks a strict correspondence between video frames and labeled sequences. After encoding the input video sequence, it is very appropriate to use CTC as a decoder. CTC was originally designed for speech recognition, mainly to perform end-to-end temporal classification of unsegmented data to solve the problem of mismatched lengths of input and output sequences. In recent years, it is often used in CSLR. CTC introduces a blank label $\{-\}$ to mark unclassified labels during decoding, that is, any word in the input video clip that does not belong to the sign language vocabulary, so that the input and output sequences can be matched, and the dynamic programming method is used for decoding\cite{li2020reinterpreting}.\par

For the input video V of T frames, the label of each frame is represented by $\pi=(\pi_1,\pi_2,...,\pi_T)$, where $\pi_t\in \nu\bigcup\{-\}$, and $\nu$ is sign language vocabulary, then the posterior probability of the label is:\par

\begin{equation}
p(\pi|V)=\prod_{\substack{t=1}}^{\substack{T}}p(\pi_t|V)=\prod_{\substack{t=1}}^{\substack{T}}Y_{t,\pi_t}
\end{equation}

For a given sentence-level label $s=(s_!,s_2,...,s_L)$, where L is the number of words in the sentence. CTC defines a many-to-one mapping B, whose operation is to remove blank labels and duplicate labels (for example, $B(-dd-og-g-)=B(-d-og-g)=dog$) in the path, then the conditional probability of label s is the sum of the occurrence probabilities of all corresponding paths:\par

\begin{equation}
p(s|V)=\sum_{\substack{\pi\in B^{-1}(s)}}p(\pi|V)
\end{equation}

Where $B^{-1}(s)=\{\pi|B(\pi)=s\}$ is the inverse mapping of B. CTC loss is defined as the negative log-likelihood of the conditional probability of s.\par

\begin{equation}
L_{CTC}=-\ln p(s|V)
\end{equation}

Then the multi-level CTC loss can be expressed as:\par

\begin{equation}
\begin{aligned}
    L_{sum}&=-\ln \prod_{\substack{i=1}}^{\substack{n}} p(s|V_i)\\
    &=-\ln (p(s|V_1)p(s|V_2)...p(s|V_{n-1})p(s|V_n))
\end{aligned}
\end{equation}

Where n is the number of CTC.\par

For the four-level gloss feature obtained after transformers, after a fully connected layer, softmax is used for normalization, and then the normalized result is decoded by CTC to obtain $L_{CTC4}$. Likewise, corresponding $L_{CTC1}$, $L_{CTC2}$, and $L_{CTC3}$ are obtained for the primary, secondary, and tertiary gloss features. Add these four CTC losses to get the final loss for training.\par

\begin{equation}
\begin{aligned}
    L_{sum}&=-\ln \prod_{\substack{i=1}}^{\substack{4}} p(s|V_i)\\
    &=-\ln (p(s|V_1)p(s|V_2)p(s|V_3)p(s|V_4))
\end{aligned}
\end{equation}

The final sign language recognition result is obtained by CTC decoding after softmax for the fourth level gloss feature only.\par

\section{Experiments}

In this section, we conduct experiments on two widely used sign language recognition datasets. We compare our model with advanced methods and perform ablation studies to demonstrate the effectiveness of each part of our model.\par

\subsection{Datasets}

The RWTH-PHOENIX-Weather-2014 (RWTH) dataset\cite{koller2015continuous}: The RWTH is recorded by a public weather radio and television station in Germany. All presenters are dressed in dark clothes and perform sign language in front of a clean background. The videos in this dataset are recorded by 9 different presenters with a total of 6841 different sign language sentences (of which the number of sign language word instances is 77321 and the number of words is 1232). All videos are preprocessed to a resolution of 210 × 260, and a frame rate of 25 frames per second (FP/S). The dataset is officially divided into 5,672 training samples, 540 validation samples, and 629 test samples.\par

Chinese Sign Language (CSL) dataset\cite{huang2018video}: The CSL is captured using a Microsoft Kinect camera and contains 100 Chinese everyday phrases, each of which is demonstrated 5 times by 50 presenters with a vocabulary size of 178. The video resolution is 1280 × 720 and the frame rate is 30 FP/S. The dataset is richer in performance diversity because the demonstrators wear different clothes and the demonstration different speed and range of motion. Without giving an official segmentation, we divide the CSL into a training set and a test set according to the 8:2 rule, with 80\% of the training set and 20\% of the test set, i.e., into a training set of 20,000 samples and a test set of 5,000 samples, and ensure that the sentences in the training and test sets are the same, but the presenters are different.\par

\subsection{Implementation details}

The model experiments in this paper are trained using the Adam optimizer, with both the initial learning rate and weight factor set to $10^{-4}$. The batch size used is 2. A total of 60 epochs are used, and the learning rate is reduced by 80\% at the 40th and 50th epochs. Data augmentation is performed using random cropping and random flipping. For random cropping, the frame size of each video sequence is first resized to $256\times 256$, and then randomly cropped to $224\times 224$ to fit the shape of the input. For random flips, its flip probability is 0.5. Flip and crop processing is performed on video sequences. We also do temporal augmentation, where the length of the video sequence grows or shrinks randomly within $\pm 20\%$. In addition, the GPU dedicated memory is limited and the amount of video data is too large. To reduce the memory footprint and improve the training speed, we use mixed precision computation with acceptable precision loss. In the final CTC decoding stage, we used a beam search algorithm for decoding with a beam width of 10. For the CSL dataset, we used 15 epochs, and the learning rate is reduced by 90\% at the 8th epoch, limited by GPU memory to down-sample the original data by a factor of 2 as input, and used only 2 levels of CTC loss. The graphics card used in this experiment is RTX2080Ti, the GPU dedicated memory size is 12G, the CPU memory is 8G, and the number of cores is 4. The proposed network model consists of three parts, each of which is detailed as follows:\par

\begin{table*}[!htbp]
\centering
\caption{Performance comparison of different continuous sign language recognition methods on the RWTH dataset}
\label{tab:aStrangeTable1}
\begin{tabular}{cccc}
\hline  
Methods& Backbone& \multicolumn{2}{c}{WER(\%)}\\
 &  &  Dev& Test\\
\hline  
Align-iOpt\cite{pu2019iterative}& 3D-ResNet& 37.1& 36.7\\
Re-Sign\cite{koller2017re}& GoogLeNet& 27.1& 26.8\\
SFL\cite{niu2020stochastic}& ResNet18& 26.2& 26.8\\
CNN+LSTM+HMM\cite{koller2019weakly}& GoogLeNet& 26.0& 26.0\\
CrossModal\cite{papastratis2020continuous}& BN-Inception& 23.9& 24.0\\
FCN\cite{cheng2020fully}& Custom& 23.7& 23.9\\
SLRGAN\cite{papastratis2021continuous}& BN-Inception& 23.7& 23.4\\
DNF\cite{cui2019deep}& GoogLeNet& 23.1& 22.9\\
CMA\cite{pu2020boosting}& GoogLeNet& 21.3& 21.9\\
VAC\cite{min2021visual}& ResNet18& 21.2& 22.3\\
STMC\cite{zhou2020spatial}& VGG11& 21.1& 20.7\\
\pmb{Our Methods}& \pmb{ResNet34}& \pmb{20.3}& \pmb{21.4}\\
\hline  
\end{tabular}
\end{table*}

\begin{table*}[!htbp]
\centering
\caption{Performance comparison of different continuous sign language recognition methods on the CSL dataset}
\label{tab:aStrangeTable2}
\begin{tabular}{cc}
\hline  
Methods& WER(\%)\\
\hline  
HLSTM\cite{guo2018hierarchical}& 7.6\\
HLSTM-attn\cite{guo2018hierarchical}& 7.1\\
Align-iOpt\cite{pu2019iterative}& 6.1\\
DPD\cite{zhou2019dynamic}& 4.7\\
SF-Net\cite{yang2019sf}& 3.8\\
FCN\cite{cheng2020fully}& 3.0\\
CrossModal\cite{papastratis2020continuous}& 2.4\\
STMC\cite{zhou2020spatial}& 2.1\\
SLRGAN\cite{papastratis2021continuous}& 2.1\\
VAC\cite{min2021visual}& 1.6\\
\pmb{Our Methods}& \pmb{0.7}\\
\hline  
\end{tabular}
\end{table*}


\begin{figure*}
  \begin{minipage}{0.5\textwidth}
\includegraphics[width=3.5in,height=2.33in]{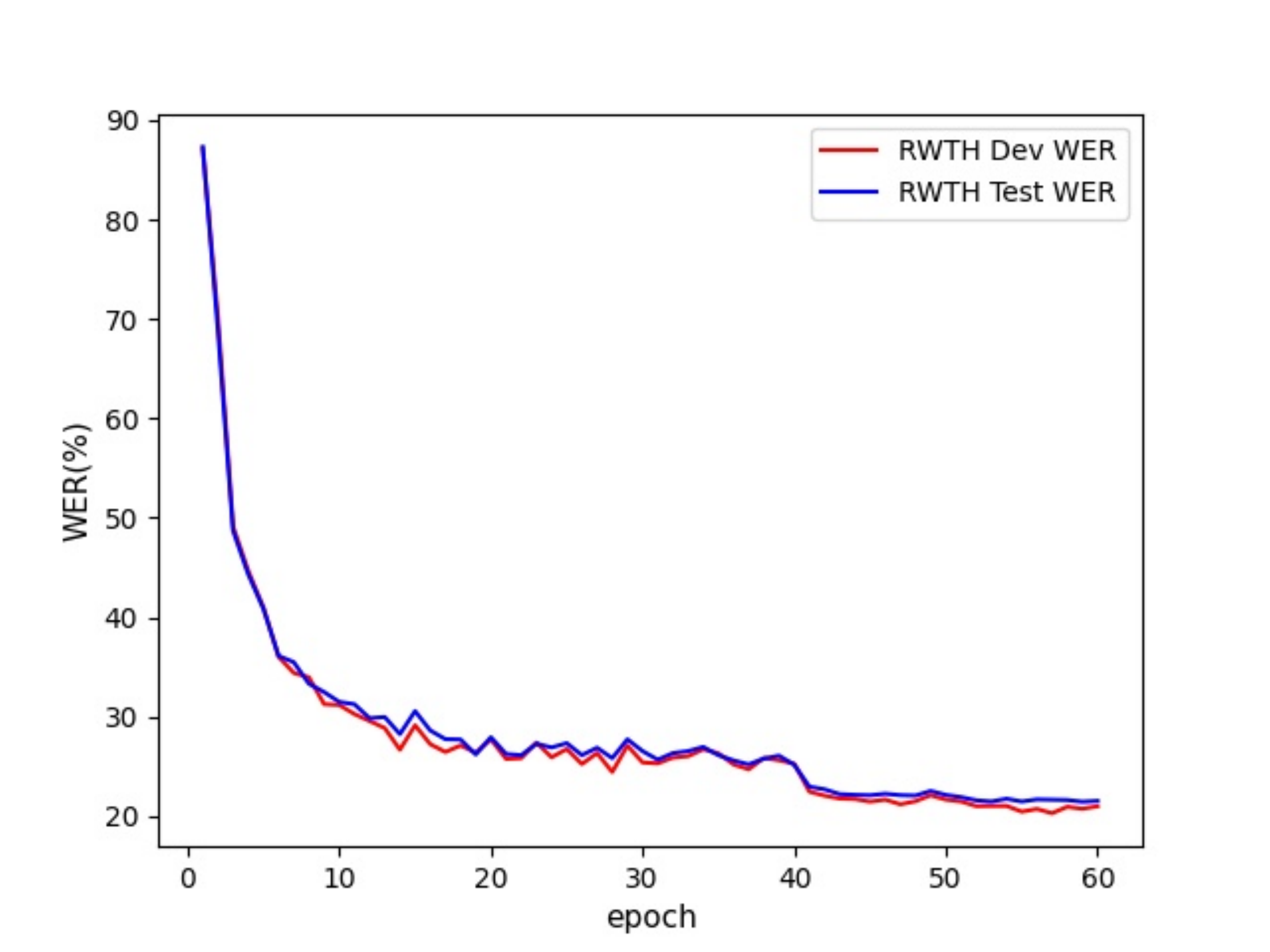}
\caption{WER variation curves for the validation and test sets of the RWTH dataset.}
\label{fig:ljxy5}
\end{minipage}
\hfill
\begin{minipage}{0.5\textwidth}
\includegraphics[width=3.5in,height=2.33in]{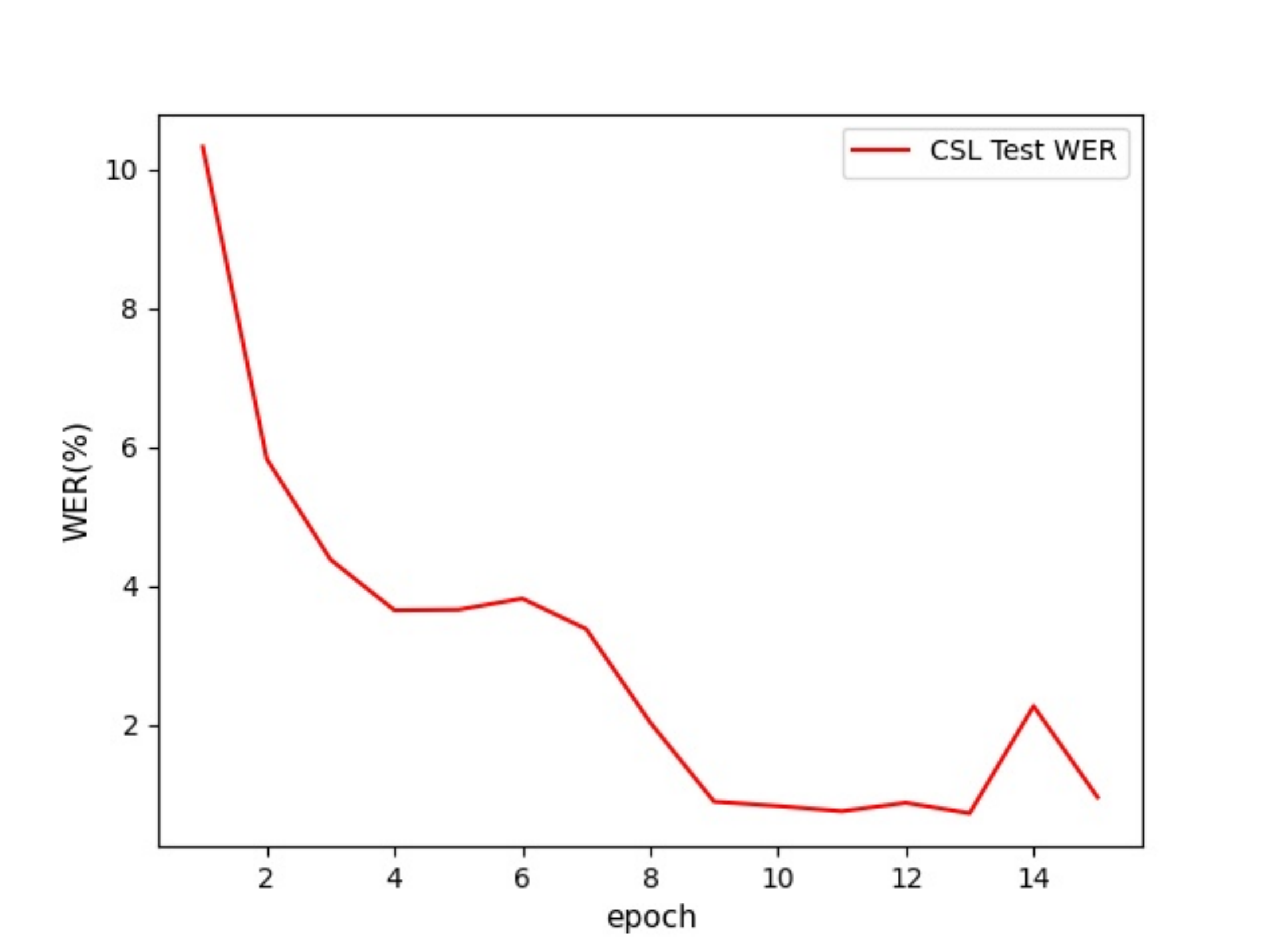}
\caption{WER variation curves for the test set of the CSL dataset.}
\label{fig:ljxy6}
\end{minipage}
\end{figure*}

Part I: Frame-wise feature extraction. We use ResNet34\cite{he2016deep} as the backbone, remove the last fully connected classification layer, and use the weight parameters trained on the ImageNet dataset\cite{russakovsky2015imagenet} as the initialization parameters. To further reduce memory usage and improve computational efficiency, a random gradient stop with a random stop probability of 0.5 is used during training. Two FC layers are connected after ResNet34 to obtain frame-wise features, and the number of channels is boosted from 512 to 1024.\par

Part II: Time-wise feature extraction. This part consists of the MST-block and the transformers encoding module. Each MST-block consists of multiple 1DCNNs with different kernels, and then the features are fused using 2DCNNs. We use 4 different kernels with kernel sizes of 3, 5, 7, and 9, padding sizes of 1, 2, 3, and 4, and stride of 1. 2DCNN has kernel = (4,2), padding = 0, and stride = 2. A total of 2 MST-blocks are used. Since the input sequence will be down-sample by a factor of 4 in this process, the size of the original video sequence needs to be converted to an integer multiple of 4 when it is input. In the transformers module, the number of input and output features is 1024, the number of layers is 2, and the number of multi-heads is 8.\par

Part III: Multi-level CTC Loss. We use 4 CTC Losses to calculate the losses of each of the 4 gloss features and sum them, and then use the summed results for training. Finally, the fourth-level of gloss features are used as decoding input for sign language recognition.\par

\subsection{Judgment Criteria}

We use the Word Error Rate (WER) to measure the performance of the model, a criterion that is widely used in CSLR\cite{koller2015continuous}\cite{guo2019hierarchical}. WER is a Levenshtein distance, which is the sum of the minimum insertion operations, substitution operations, and deletion operations required to convert a recognition sequence into a standard reference sequence. Lower WER means better recognition performance, and its definition is as follows:\par

\begin{equation}
WER=100\%\times \frac{ins+del+sub}{sum}
\end{equation}

Where “ins” represents the number of words to be inserted, “del” represents the number of words to be deleted, “sub” represents the number of words to be replaced, and “sum” represents the total number of words in the label. Because “ins+del+sub” may be greater than sum, the result of WER may be greater than 100\%. When experimenting with the CSL dataset, we treat a single Chinese character as a word.\par

\begin{figure*}
  \begin{center}
  \includegraphics[width=5in]{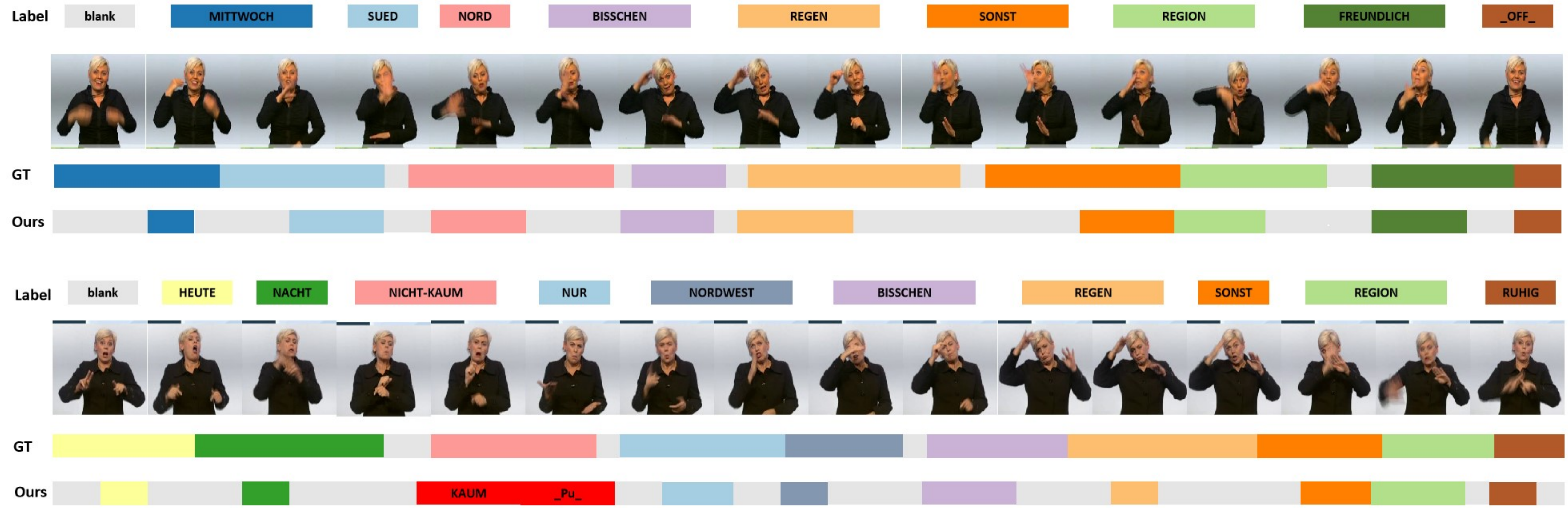}\\
  \caption{Two sample recognition results on the RWTH validation set.}\label{fig:ljxy7}
  \end{center}
\end{figure*}

\begin{table*}[!htbp]
\centering
\caption{Experimental results of different numbers of 1DCNNs in MST-block on RWTH}
\label{tab:aStrangeTable3}
\begin{tabular}{ccc}
\hline  
kernel& Dev(\%)& Test(\%)\\
\hline  
1& 21.5& 22.8\\
2& 21.4& 22.0\\
3& 20.7& 21.0\\
\pmb{4}& \pmb{20.3}& \pmb{21.4}\\
5& 20.4& 21.6\\
\hline  
\end{tabular}
\end{table*}

\begin{table*}[!htbp]
\centering
\caption{Experimental results of adding different number of FC layers on RWTH after Backbone}
\label{tab:aStrangeTable4}
\begin{tabular}{ccc}
\hline  
The number of FC layers& Dev(\%)& Test(\%)\\
\hline  
0& 21.0& 21.5\\
1& 20.8& 21.1\\
\pmb{2}& \pmb{20.3}& \pmb{21.4}\\
3& 21.2& 21.8\\
\hline  
\end{tabular}
\end{table*}

\subsection{Experimental results}

The experimental results on the RWTH dataset and the CSL dataset are shown in Table \RNum{1} and Table \RNum{2}, respectively. The WER curves generated for each epoch in the training process are shown in Figures 5 and 6.\par

As can be seen from the Table \RNum{1} and \RNum{2}, the proposed model achieves better performance compared to other advanced models on both datasets. For the RWTH dataset, our model achieves 20.3\% and 21.4\% on the validation and test sets, respectively. On the CSL dataset, the best result reaches 0.7\%. The effectiveness of the proposed model is demonstrated. It can be seen from the curve that WER decreases with the increase of epoch. WER decreases significantly when the first Learning rate changes. RWTH reaches the minimum value at the 57th epoch, and CSL reaches the minimum value at the 13th epoch.\par

To further demonstrate the effectiveness of the proposed model, we present some sample outputs in Figure 7, showing the recognition results on the RWTH validation set. Where “Label” represents the real sentence corresponding to the sequence, “GT” represents the correspondence between the real sentence and the sequence, the correspondence between the sentence and the sequence in “GT” is the result of manual alignment, “Ours” represents the recognition result of the model in this paper, and “red” represents the wrong one recognition. The video frame sequence in Figure 7 is the result of down-sampling the original video frame sequence. It can be seen from Figure 7: for the first sample, the sentence can be accurately recognized; in the second sample, the third word is misidentified from "NICHT-KAUM" to "KAUM", and the word "\_Pu\_" is misidentified between the third and fourth words.\par

\subsection{Ablation studies}

In this section, we conduct ablation experiments on the RWTH dataset to further verify the effectiveness of the model. All ablation experiments use WER as the metric.\par

1) Experiment 1: The number of 1DCNNs in MST-block. The MST-block proposed in this paper utilizes different scales of temporal receptive fields to improve the ability of sequence modeling. It mainly consists of multiple 1DCNNs with different convolution kernels in parallel to form a multi-scale feature extraction part. The influence of the number of 1DCNNs used on the experimental results is shown in Table \RNum{3}. The kernel size of the starting 1DCNN is 3, and each additional 1DCNN increases the kernel size by 2.\par

It can be seen that with the increase of the number of 1DCNNs, the WER in the table has a tendency to decrease first and then increase. When the number of 1DCNNs increases to 4, the recognition effect is the best, and the WER decreases to 20.3\%, and the WER starts to rise slightly when it increases to 5 1DCNNs.\par

2) Experiment 2: The number of FC layers used after Backbone. In this paper, the function of adding two FC layers after Backbone is to integrate the features in the image feature map after multiple convolution layers and pooling layers to obtain the high-level meaning of image features. The influence of the number of FC layers used on the experimental results is shown in Table \RNum{4}.\par

It can be seen that the value of WER is 21.0\% when no FC layer is added. As the number of FC layers increases, the WER first decreases and then increases. When the number of FC layers is 2, it reaches a minimum value of 20.3\%.\par

3) Experiment 3: Enhancement coding method. In this paper, for the temporal feature vector obtained by MST-block, the temporal features are further encoded using transformers encoding module, so as to make the obtained temporal features more accurate. For the effect of using different enhanced coding methods on the experimental results, it is shown in Table \RNum{5}. We use BiLstm and transformers for further encoding of temporal features for comparison.\par

\begin{table*}[!htbp]
\centering
\caption{Experimental results of different enhancement coding methods on RWTH}
\label{tab:aStrangeTable5}
\begin{tabular}{ccc}
\hline  
Enhancement coding method& Dev(\%)& Test(\%)\\
\hline  
BiLstm& 21.3& 21.7\\
\pmb{Transformers}& \pmb{20.3}& \pmb{21.4}\\
\hline  
\end{tabular}
\end{table*}

\begin{table*}[!htbp]
\centering
\caption{Experimental results of different numbers of CTC losses on RWTH}
\label{tab:aStrangeTable6}
\begin{tabular}{ccc}
\hline  
The number of CTC losses& Dev(\%)& Test(\%)\\
\hline  
1& 28.1& 28.4\\
2& 22.1& 23.4\\
3& 21.0& 22.1\\
\pmb{4}& \pmb{20.3}& \pmb{21.4}\\
\hline  
\end{tabular}
\end{table*}

It can be seen that using the transformers module can better encode the temporal features compared to BiLstm. BiLstm is widely used for temporal feature modeling in CSLR, and it works well. Here, the reason why BiLstm is not as good as transformers is that when further encoding the temporal features extracted by MST-block, it is necessary to pay more attention to global features, while BiLstm pays more attention to local features.\par

4) Experiment 4: The number of CTC losses for multi-level CTC loss. In this paper, to address the problem that the deep neural network uses the backward propagation algorithm based on the chain rule to train the network, which to a certain extent causes the shallow network parameters to not be well learned and updated, multi-level CTC loss is proposed to decode the temporal features, and then efficiently train the feature extraction network and temporal modeling network to further improve the recognition performance. For the effect of using the number of CTC losses on the experimental results, it is shown in Table \RNum{6}.\par

Levels 1 to 4 CTC correspond to levels 1 to 4 gloss features, and when performing multi-level CTC ablation experiments, the retention of CTC levels is in reverse order. That is, when one CTC is used, the last CTC level is retained, and when two CTCs are used, the last two CTC levels are retained. When only two CTCs are used in the ablation experiments, then the used is the sum. As can be seen in Table \RNum{6}., the WER is in a downward trend as the number of multi-level CTCs increases.\par

\section{Conclusion}

This paper proposes a novel MSTNet for CSLR. In this work, to address the lack of accurate annotation of data time series in CSLR, we propose a multi-scale temporal receptive field approach to obtain more accurate temporal features and thus improve the accuracy of CSLR. The resulting temporal features will be more accurate compared to the method of extracting temporal features using a fixed temporal receptive field in recent research work\cite{cui2019deep}\cite{min2021visual}\cite{cheng2020fully} on CSLR. It will also be richer and more comprehensive than the method in\cite{wei2020semantic} that generates temporal features of different scales by using different strides of pooling for temporal features. Furthermore, our proposed multi-level CTC loss is able to better learn and update the shallow network parameters in the CNN, effectively train the network and further improve the model recognition performance. This is superior to the multi-scale perceptual loss used in\cite{wei2020semantic} because their proposed multi-scale perceptual loss is not helpful for the training of shallow networks. In our proposed model MSTNet, the temporal feature extraction part first extracts temporal receptive field features of different scales through the proposed MST-block, and combines them into a candidate space, which is fused through learnable parameters, greatly improving the temporal modeling capability, and then encodes them by the transformers module to better obtain remote dependence information, which further improves the accuracy of the final temporal features. Finally, training with the proposed multi-level CTC loss improves the accuracy of sign language recognition. The entire network is trained end-to-end and experimentally validated on two large-scale sign language datasets, and the experimental results demonstrate the effectiveness of the model proposed in this paper.\par

We believe that a possible future research direction for CSLR is to keep the final recognition rate essentially constant after down-sampling the temporal dimension of the input sign language video data. There have been some similar studies, such as video super-resolution\cite{chan2021understanding}\cite{liu2022video}, but there is no such study in continuous sign language recognition yet. The sign language actions in the video are continuous in time, and using fewer video frames to describe the sign language actions can reduce redundancy, which in turn can improve network execution efficiency and reduce memory usage. Therefore, down-sampling in the temporal dimension is an effective way to improve the real-time performance of CSLR. However, down-sampling will lead to accuracy degradation. How to keep the accuracy basically unchanged while down-sampling is a problem worthy of study.\par

\section*{Acknowledgment}

This work was supported in part by the Development Project of Ship Situational Intelligent Awareness System, China under Grant MC-201920-X01, in part by the National Natural Science Foundation of China under Grant 61673129. \par


%





\ifCLASSOPTIONcaptionsoff
  \newpage
\fi





\bibliographystyle{IEEEtran}
\bibliography{IEEEabrv,Bibliography}

\vfill


\end{document}